# BCOT: A Markerless High-Precision 3D Object Tracking Benchmark

Jiachen Li[1], Bin Wang[1], Shiqiang Zhu[2], Xin Cao[1], Fan Zhong[1], Wenxuan Chen[2],
Te Li[2*], Jason Gu[3] and Xueying Qin[1*]

[1]Shandong University   [2]Zhejiang Lab   [3]Dalhousie University

## Abstract

*Template-based 3D object tracking still lacks a high-precision benchmark of real scenes due to the difficulty of annotating the accurate 3D poses of real moving video objects without using markers. In this paper, we present a multi-view approach to estimate the accurate 3D poses of real moving objects, and then use binocular data to construct a new benchmark for monocular textureless 3D object tracking. The proposed method requires no markers, and the cameras only need to be synchronous, relatively fixed as cross-view and calibrated. Based on our object-centered model, we jointly optimize the object pose by minimizing shape re-projection constraints in all views, which greatly improves the accuracy compared with the single-view approach, and is even more accurate than the depth-based method. Our new benchmark dataset contains 20 textureless objects, 22 scenes, 404 video sequences and 126K images captured in real scenes. The annotation error is guaranteed to be less than 2mm, according to both theoretical analysis and validation experiments. We re-evaluate the state-of-the-art 3D object tracking methods with our dataset, reporting their performance ranking in real scenes. Our BCOT benchmark and code can be found at https://ar3dv.github.io/BCOT-Benchmark/.*

## 1. Introduction

Template-based 3D object tracking aims to estimate the accurate 6DOF pose of moving objects with known 3D models. It is an essential task of computer vision [21], and is widely used in applications that desire high-precision 3D object pose, such as augmented reality, robotic grasping, etc. Despite the rapid development of single-frame 6DOF pose estimation methods [32, 41], for video analysis 3D tracking can be more accurate and more efficient, and thus is indispensable.

Because it is difficult to annotate the accurate 3D pose of a moving object in the real video, it is a great challenge to evaluate 3D tracking methods in real scenes. Previous works use only synthetic datasets or datasets with low precision, the movement of object and camera is also limited. Currently the main adopted datasets include RBOT [38], OPT [45] and YCB-Video [47]. The RBOT dataset is semi-synthetic with rendered moving objects in the real background image sequences, which may be different from real video in camera effects and object movement. The OPT dataset is real captured, but with numerous artificial markers around the object, and the objects are not allowed to move. The YCB-Video is a real RGB-D dataset without markers; however, it contains only static objects, and the pose annotation contains a significant error that prevents it from being used in high-precision scenarios (as required by many AR applications). The above limitations are especially important for the learning-based tracking methods [43, 44], due to the domain difference as well as the bias to the visual cues and object movements in the training dataset.

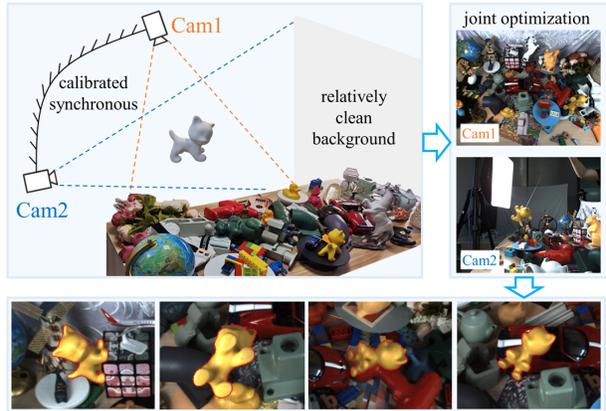

Figure 1. The capturing camera set is composed of two approximately orthogonal cameras. Based on the binocular data and proposed joint optimization framework, we can organize the benchmark with precise annotated pose as rendered in red contour.

Table 1 lists and compares related datasets, including some datasets for single-frame pose estimation [1, 13, 14, 18, 47]. Note that even with the depth camera, it is still difficult to annotate the accurate 3D pose, due to the depth er-ror and mis-alignment with RGB image, especially around the object boundary [47]. On the other hand, the datasets with artificial markers [1, 13, 14, 18], will not only damage the naturalness of the scene, but also limit the object movement. Therefore, a high-precision markerless pose estimation method is required in order to build a real-life 3D tracking benchmark, which according to our knowledge has not been addressed in previous works.

| Dataset | data type | marker-less | depth | outdoor | dynamic object | for tracking | objects | sequences | frames |
| --- | --- | --- | --- | --- | --- | --- | --- | --- | --- |
| Linemod [13] | real | × | ✓ | × | × | × | 13 | - | 15K |
| Linemod Occlusion [1] | real | × | ✓ | × | × | × | 8 | - | 1.2K |
| T-LESS [14] | real | × | ✓ | × | × | × | 30 | - | 49K |
| HomebrewedDB [18] | real | × | ✓ | × | × | × | 33 | - | 17K |
| TOD [27] | real | × | ✓ | × | × | × | 20 | - | 64K |
| StereOBJ-1M [26] | real | × | × | ✓ | × | × | 18 | - | 397K |
| YCB-Video [47] | real | ✓ | ✓ | × | × | ✓ | 21 | - | 134K |
| OPT [45] | real | × | ✓ | × | × | ✓ | 6 | 552 | 101K |
| RBOT [38] | semi-synthetic | ✓ | × | × | ✓ | ✓ | 18 | 72 | 72K |
| **BCOT (Ours)** | real | ✓ | × | ✓ | ✓ | ✓ | 20 | 404 | 126K |

Table 1. Dataset Comparisons. Our BCOT benchmark is the only real scene benchmark that provides dynamic objects. The benchmark doesn't have invasive artificial markers and includes indoor and outdoor scenes.

In this paper we propose a markerless multi-view approach to address the above problem. Our method can estimate the high-precision 3D pose of video objects from two orthogonal views of RGB images captured by the high-resolution, high-speed and synchronous cameras, as illustrated in Fig. 1. To deal with textureless objects, we adopted a shape-based method, which does not rely on point correspondences between views. The object pose is solved by jointly optimizing the shape re-projection error of multiple views, in a novel object-centered pose estimation framework. Based on the proposed approach we contribute a new 3D tracking benchmark, namely **BCOT (BinoCular Object Tracking) benchmark**, which contains accurately annotated real videos, with both camera and objects can move freely. The maximum annotation error of the BCOT benchmark is $2mm$, achieving the best annotation precision at present. The major contributions can be summarized as follows.

- We propose a multi-view approach that can estimate the accurate 3D pose of real video objects. Our method is markerless and suitable for textureless and moving objects, and thus provides a way to annotate real-life tracking videos.

- We build a 3D tracking benchmark of real scenes with high-precision ground truth (GT) poses, with annotation accuracy guaranteed by both theoretical analysis and validation experiments.

- We comprehensively evaluate the existing SOTA 3D object tracking methods on the proposed BCOT benchmark.

## 2. Related Work

**Related datasets.** Early 3D tracking algorithms usually use their own collected video sequences as test data [3, 31, 33, 37, 39, 42], which scale is small and difficult to reflect the performance of the algorithm.

In recent years, some large-scale 3D object tracking datasets have been released. The RBOT dataset [38] is a semi-synthetic dataset with the real scene background and the rendered object. It provides absolute GT poses but lacks authenticity. The motion trajectory of the virtual object is also pre-set, and all sequences share the same trajectory, limiting the diversity of object motion. The OPT dataset [45] is a real scene dataset, which uses artificial markers to calculate the GT pose. However, the marker occupies most background areas, making the background invasive.

For pose estimation datasets based on the single frame [1, 6, 8, 13, 14, 18, 26, 27, 47], because they do not annotate the pose on the sequence, or the annotation precision is insufficient, they cannot effectively evaluate the 3D tracking method.

**Monocular textureless 3D object tracking.** Textureless 3D object tracking can be divided into edge-based methods [2, 5, 9, 29, 34, 39, 40, 42, 46] and region-based methods [11, 33, 35–38, 49] according to the feature used. There are also some methods based on feature fusion, which use multiple features explicitly [23, 24, 48] or implicitly [15, 16, 42] to achieve better results.

In recent years, some methods based on deep learning have been explored [4, 7, 25, 43, 44], but their performance is still not comparable to those based on traditional features.

**Multi-view tracking.** Multi-view geometry [10] is widely used in computer vision, which estimates the camera and object pose through corresponding feature matching [13, 14, 17, 28]. However, the constraints will fail for textureless objects because the stable point or edge features cannot be extracted and matched when the object lacks the texture.

Li et al. [22] and Labbé et al. [20] propose a two-step multi-view optimization framework for textureless objects. They first estimate the object pose using the image feature under every single camera and then minimize a reconstruction loss (irrelevant to image feature) in the world coordinate frame, where [20] uses 2D projection points and [22] uses 3D points. This kind of strategy separates the features and coordinate system, limiting the precision. Besides, they are single frame pose estimation methods whose accuracy is lower than the tracking method.

## 3. Multi-view Pose Estimation

Since the single-view RGB-based method can make use of only 2D reprojection error for 3D pose estimation, it is error-prone in the direction of the camera view ($Z$-axis), as shown in Fig. 2. In multi-view tracking, each camera has a different view direction, and the uncertainty thus can be greatly reduced.

### 3.1. The Object-centered Model

We should note that all previous 3D tracking methods are based on the camera coordinate frame to optimize the object pose, which can not directly establish the association between multiple cameras. So we first select a basic coordinate frame. Considering that all cameras are directly associated with objects, we use the object-centered coordinate frame $O_o$ for pose optimization, which takes the center of the object model as the coordinate origin, as shown in Fig. 2. Please note that object template coordinate frame $O_t$ is different from $O_o$. The origin of $O_t$ can be at any position, which is determined by the CAD model. At the same time, we need to know the relative position between the cameras. Here, we call the model based on the camera coordinate frame $O_c$ the camera-centered model, and the model based on the object-centered coordinate frame $O_o$ is called the object-centered model.

When using multi-view information, the coordinate frames between multiple cameras are no longer independent. So the pose cannot be estimated independently in each camera coordinate frame $O_{c_i}$ but should be solved jointly based on the basic coordinate frame $O_o$. The subscript $i$ represents the camera index. We need to re-derive the camera projection model and pose updating process. The projection model in $O_o$ is formulated as:

$$x = \pi(K({}^cT_t \tilde{X}_t)_{3\times 1}) \quad (1)$$
$$= \pi(K({}^oT_c^{-1}\,{}^oT_c\,{}^cT_t \tilde{X}_t)_{3\times 1}) \quad (2)$$
$$= \pi(K({}^oT_c^{-1} \tilde{X}_o)_{3\times 1}). \quad (3)$$

$X_m$ represents the 3D point of the object, and the subscript $m$ represents the corresponding coordinate frame $O_m$. $\tilde{X} = (X, Y, Z, 1)^\top$ is homogeneous representation of the $X = (X, Y, Z)^\top = (\tilde{X})_{3\times 1}$ and $\pi(X) = [X/Z, Y/Z]^\top$.

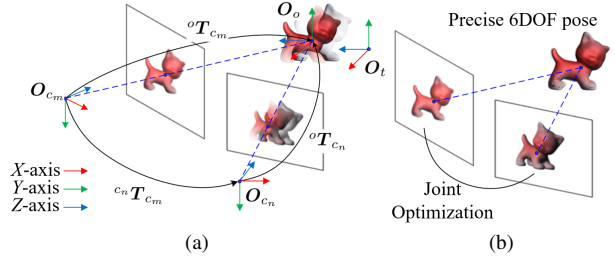

Figure 2. Error analysis and elimination of monocular 3D tracking. (a) When only using the left view for optimization, the predicted 3D position (in gray) is very different from that of the GT position (in red), even if there is a plausible visual result. This means there is a large translation error along the viewing direction, where the re-projection error can be observed in the right view. (b) Through the proposed joint optimization framework, we can obtain the precise 6DOF pose, which can render the exact contour on each image.

$x$ is the 2D point on the image and $K$ is the pre-calibrated camera intrinsic parameter. ${}^nT_m$ represents the coordinate transformation from $O_m$ to $O_n$. In particular, ${}^cT_t$ also represents the object pose in $O_c$. ${}^oT_c$ can be calculated from the $O_t, O_o$ and the initial pose (i.e., the pose of the previous frame). In practical applications, it can also select $O_t, O_{c_i}$, or other unified coordinate frames.

### 3.2. Joint Pose Estimation

The monocular energy function in $O_o$ is formulated as:

$$\Delta\boldsymbol{\xi}_o = \arg\min_{\Delta\boldsymbol{\xi}_o} \sum_{\boldsymbol{x}\in\Omega} F(\boldsymbol{x}, \boldsymbol{\xi}'_c, {}^oT_c), \quad (4)$$

and we then map the updated pose increment $\Delta\boldsymbol{\xi}_o$ to $O_c$ by $\Delta T_c = {}^cT_o exp(\Delta\hat{\boldsymbol{\xi}}_o){}^cT_o^{-1}$. $\boldsymbol{\xi}'_c$ is the initial pose in $O_c$. $F$ represents the arbitrary energy function for 3D tracking, here we adopt the region-based pose estimation method introduced in [38]. $F$ in $O_o$ can be re-formulated as:

$$F(\boldsymbol{x}, \boldsymbol{\xi}'_c, {}^oT_c) = F(\boldsymbol{x}, \boldsymbol{\xi}_o) \quad (5)$$
$$= -\log\big(H_e(\Phi(\boldsymbol{x}(\boldsymbol{\xi}_o)))P_f(\boldsymbol{x}) + (1 - H_e(\Phi(\boldsymbol{x}(\boldsymbol{\xi}_o))))P_b(\boldsymbol{x})\big), \quad (6)$$

which minimizes the shape reprojection error with respect to a rendered shape template $\Phi$ and estimated soft object segmentation $P_f, P_b$. The original $F$ in method [38] is associated with the pose $\boldsymbol{\xi}_c$ in $O_c$, and we use the object-centered model to associate $F$ with $\boldsymbol{\xi}_o$, and solve for the pose increment in $O_o$. For the case of multiple views, we propose to optimize the following joint energy function:

$$\Delta\boldsymbol{\xi}_o = \arg\min_{\Delta\boldsymbol{\xi}_o} \sum_{i=1}^{N} \sum_{\boldsymbol{x}\in\Omega^i} F^i(\boldsymbol{x}, \boldsymbol{\xi}'_{c_i}, {}^oT_{c_i}) \quad (7)$$
$$= \arg\min_{\Delta\boldsymbol{\xi}_o} \sum_{\boldsymbol{x}\in\Omega} F(\boldsymbol{x}, \boldsymbol{\xi}_o), \quad (8)$$

in which $N$ represents the number of views. $\Omega^i$ represents optimized points set in each view. Through the object-centered model, we combine all the optimized points in each view to form an optimized area $\Omega$, where all sample points are independent. Eq. 7 uses maximum likelihood estimation to maximize the color probability difference between the foreground and background to solve the pose. It becomes a summation by taking the logarithm. Therefore, all sampling points can be jointed, and Eq. 8 can constrain the pose by unifying the image features and coordinate system.

During tracking, the object can freely move. The camera movement needs to be discussed in two cases. The first case is that the relative spatial position of the cameras is fixed, and the transformation between the cameras can be calibrated in advance. The second case is that the cameras move freely. At this time, we need to calibrate the camera in real-time during the tracking. For example, put an artificial marker [19] in the scene or use SLAM [30] or other technologies, but this way will introduce the calibration error. For data collection we prefer the first approach, which allows the cameras to move with fixed relative pose (see section 4.3).

*Optimization*. We use the Gauss-Newton method to solve the joint energy function. Please refer to the supplementary materials for details.

## 4. BCOT Benchmark

Based on our high-precision multi-view tracking method, we construct the BCOT benchmark.

### 4.1. Data Collection

We utilize two high-resolution, high-speed cameras (MER-131-210U3) to capture images synchronously in the data collection stage, using Cam1 and Cam2 to indicate. The camera exposure time is $5ms$ ($200FPS$) so that no motion blur will occur. The image resolution is $1280 \times 1024$. When storing data, limited by the transmission bandwidth of the USB3.0 interface, the images of the two cameras can only be stored at a speed of $60FPS$, but the storing speed doesn't influence the exposure time. The included angle between the two cameras is approximately $90°$. Their relative position in the scene is fixed, and we pre-calibrate them in advance. During collection, the camera set and the object can move freely in the scene.

We then use the binocular images taken by the Cam1 and Cam2 cameras to annotate the object pose. As shown in Fig. 1, the image taken by Cam1 is the complex scene designed in advance, and the image taken by Cam2 is the relatively clean background to improve the annotation precision. The pose annotation process does not need to consider the calculation efficiency. We increase the number of iterations in the optimization process to ensure convergence. The initial pose of the first frame can be roughly set manually. With the help of the spatial position relationship constraints between the cameras, it can be optimized to a precise value. We annotate the object pose frame by frame under the high frame rate ($60FPS$) image. At this time, the pose increment between frames is smaller, which can obtain higher precision. After annotating the entire sequence, we downsample the frames and resolutions, i.e., $30FPS$, $640 \times 512$ resolution. This strategy will annotate a more precise object pose.

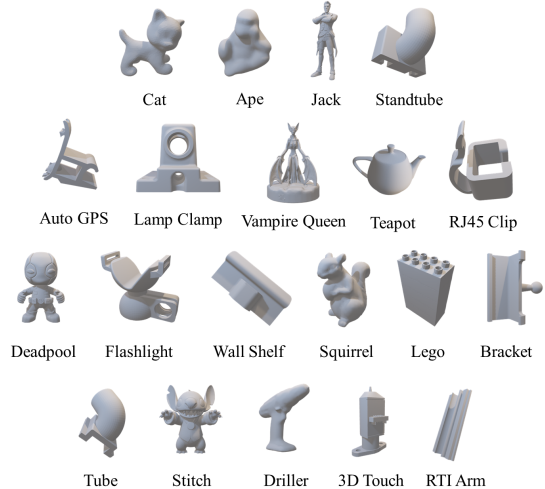

Figure 3. 3D Models in the BCOT Benchmark.

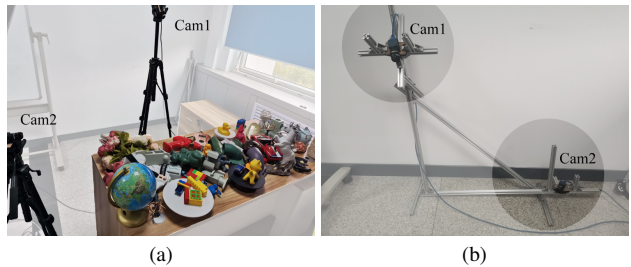

Figure 4. Camera layout: (a) mounted on tripods; (b) mounted on a movable bracket.

Finally, the images with annotated object poses are provided to users. In this way, we can collect images of complex scenes and provide precise poses of objects without any invasion of artificial markers, ensuring the authenticity of the scene and the movement of the object.

### 4.2. 3D Models

The BCOT benchmark contains 20 objects, as shown in Fig. 3. The first row is the irregular object, and the second row is the hollow object. The last three rows are symmetrical objects. The real object is 3D printed according to the model and painted with one single color to ensure texture-less. Besides, all objects have reflective properties. The range of the longest side of the model bounding box is from $91.7mm$ to $229.5mm$.

### 4.3. Scenes

The BCOT benchmark contains a variety of scene attributes and multiple motion modes, which are partially combined to a total of 22 scenes.

**Static camera set.** The camera set in the scene are fixed, so its background is basically static, with only a few objects moving slowly on the turntable. Objects move freely in the scene. The camera layout is shown in Fig. 4(a).

**Movable camera set.** The camera set are fixed on a bracket, as shown in Fig. 4(b). The relative pose between the cameras is fixed, and the camera set can be moved freely. The other configurations are the same as the static camera set. When moving the bracket, the camera on it may shake very slightly. Therefore, we will slowly move the bracket to ensure annotating precision.

**Indoor scenes.** For the indoor scene, we only provide users with images taken by Cam1, where the background is delicately designed, as shown in Fig. 1. At the same time, we construct simple scenes and complex scenes, respectively.

**Outdoor scenes.** For the outdoor scene, the images taken by the two cameras have the same priority, so we provide the images taken by two cameras to the user.

**Motion modes.** The object motion is mainly divided into three modes. 1) Translation movement: tie the object to the toy car while manipulating the car to produce the free translation. 2) Suspension movement: use a transparent fishing line to tie the object to perform suspension movement. At the same time, move the apex to generate random motion and rotation. 3) Handheld movement: people hold the object to move freely.

**Dynamic light.** In indoor scenes, we added dynamic lighting sources to increase the complexity of the scene. In outdoor scenes, natural light at different times will naturally produce light changes.

**Occlusion.** In the suspension movement mode, we tie two objects at the same time to create mutual occlusion. Because occlusion will impact the precision of our multi-view tracking method, there are fewer occlusion sequences in the BCOT benchmark.

### 4.4. Data Post-processing

Some models may face considerable challenges in some specific scenes, e.g., reflective, symmetrical, and fast-rotating scenes, where our joint optimization framework cannot precisely track objects. If the reprojection error on the original resolution image is greater than 2 pixels, we will discard the sequence. In the BCOT benchmark, we provide 404 valid sequences.

| Dataset | RGB-D datasets | TOD [27] | StereOBJ-1M [26] | BCOT |
|---------|----------------|----------|------------------|------|
| Error   | $\geq 17mm$    | $3.4mm$  | $2.3mm$          | **<2.0***mm* |

Table 2. Annotation error in 3D space.

### 4.5. Why Use Binocular Data?

The critical factor determining our tracking precision is the included angle between the cameras (see section 5.1). The orthogonal angle can already constrain the object in space and eliminate the uncertainty of the pose. New uncertainties may be introduced to affect the precision if other cameras are added in between the orthogonal cameras.

If the cameras are arranged uniformly in the space under various viewing, objects can be constrained to the greatest extent. But it will also bring storage and camera movement restrictions, and camera synchronization will also introduce extra errors. Therefore, to balance precision and operability, we adopted two cameras.

### 4.6. Error Analysis

The error of pose annotation mainly comes from three aspects: calibration error between cameras, camera synchronization, and estimation error of the proposed multi-view tracking method. The calibration between the cameras is an offline process, so it can be considered precise. We then use the system clock to ensure the synchronization between cameras.

We comprehensively analyze the errors caused by camera synchronization and pose estimation. In the binocular data, we analyze the annotation error by observing the re-projection error of the object contour. Cameras with 90° included angles can constrain objects from 3 directions, one camera constrains the $X$ and $Y$ axes, and the other camera constrains the $Z$-axis. Ideally, the projection contour under the annotation pose can be precisely aligned with the object contour in each image.

The reprojection pixel error can be observed from the binocular data. We then use the camera projection model in Eq. 1 to convert to spatial position error. In the formula, $^c\boldsymbol{T}_t$ is the annotated object pose, which converts the object template coordinate frame $\boldsymbol{O}_t$ to the camera coordinate frame $\boldsymbol{O}_c$. $^c\boldsymbol{T}_t$ is the rigid transformation matrix, and $\tilde{\boldsymbol{X}}_t$ is the model vertices, so the error of $\tilde{\boldsymbol{X}}_c$ in the camera coordinate frame is equivalent to the annotation error $^c\boldsymbol{T}_t$. We then expand the formula to get the mapping relationship:

$$x = \frac{f_x}{Z_c}X_c + c_x, \quad y = \frac{f_y}{Z_c}Y_c + c_y. \quad (9)$$

$c_x$ and $c_y$ are constants, so $X_c$ and $Y_c$ are proportional to $x$ and $y$, respectively. In the original data of our benchmark, $\frac{f_x}{Z_c}$ and $\frac{f_y}{Z_c}$ are both in $[1, 2]$, which means each pixel error corresponds to a spatial 3D error of $0.5-1mm$. In all sequences, we ensure that the pixel errors in two views are both within 2 pixels, so the maximum spatial error of the proposed benchmark is $2mm$, which is the benchmark with the highest annotation precision at present. Table 2 shows the annotation error compared to the other datasets. The RGB-D sensor has a random error standard deviation of $17mm$[1], and the keypoints-based annotation methods [27] and [26] have the RMSE (Root Mean Squared Error) of $3.4mm$ and $2.3mm$.

| Included Angle | Mono. | 5° | 10° | 20° | 30° | 45° | 60° | 90° | 120° | 5° | 10° | 30° | 45° |
|---|---|---|---|---|---|---|---|---|---|---|---|---|---|
| Camera Index | C-0 | C-0/C-1 | C-0/C-2 | C-0/C-3 | C-0/C-4 | C-0/C-5 | C-0/C-6 | C-0/C-7 | C-0/C-8 | C-0/C-9 | C-0/C-10 | C-0/C-11 | C-0/C-12 |
| **r**(°) | 1.62 | 1.31 | 1.22 | 1.17 | 1.07 | 0.94 | 0.87 | 0.76 | **0.62** | 1.33 | 1.27 | 1.06 | 0.82 |
| tx(mm) | 4.36 | 3.27 | 2.12 | 1.12 | 0.80 | 0.57 | 0.41 | **0.28** | 0.31 | 3.18 | 2.01 | 0.53 | 0.40 |
| ty(mm) | 2.39 | 1.80 | 1.15 | 0.55 | 0.34 | 0.28 | 0.26 | 0.23 | **0.22** | 1.82 | 1.19 | 0.45 | 0.38 |
| **tz**(mm) | 22.09 | 16.67 | 10.64 | 5.11 | 2.86 | 1.35 | 0.67 | **0.28** | 0.37 | 16.30 | 10.48 | 1.72 | 0.93 |
| Lost Number | 21 | 15 | 10 | 4 | 1 | 0 | 0 | 0 | 0 | 18 | 11 | 0 | 0 |

Table 3. Binocular tracking evaluation on *Object moves freely with fixed cameras* mode.

### 4.7. Evaluation Metric

We use $n°$, $n\ cm$ and ADD metric to evaluate the monocular tracking method.

**n°**, **n** $cm$. The tracking is considered accurate when the rotation error is less than $n°$ and the translation error is less than $n\ cm$ [38]. This value is usually set to $5°$, $5cm$. If it is greater than this pre-set value, the initial pose will be reset to the GT pose. For indoor objects, $5cm$ is usually a large threshold, so we will adjust the value of $n$ to re-evaluate the monocular 3D tracking method. Given the GT translation $\boldsymbol{t}$ and rotation $\boldsymbol{R}$, and the predicted translation $\hat{\boldsymbol{t}}$ and rotation $\hat{\boldsymbol{R}}$, the translation error and rotation error are defined as:

$$e(\boldsymbol{t}) = \|\hat{\boldsymbol{t}} - \boldsymbol{t}\|_2, \quad (10)$$

$$e(\boldsymbol{R}) = \cos^{-1}\left(\frac{1}{2}(\text{trace}\left(\hat{\boldsymbol{R}}^\top \boldsymbol{R}\right) - 1)\right). \quad (11)$$

**ADD metric**. ADD metric [12] represents the average distance between the model points in the predicted pose and the GT pose. When the error is less than the preset value, the tracking is considered correct. Although our BCOT benchmark contains many symmetrical objects, we have not used the ADD-S metric for symmetrical objects because we have the pose of the previous frame as the prior information. The ADD metric is formulated as:

$$ADD = \frac{1}{M}\sum_{i=1}^{M} \|(\hat{\boldsymbol{R}}\boldsymbol{X} + \hat{\boldsymbol{t}}) - (\boldsymbol{R}\boldsymbol{X} + \boldsymbol{t})\|. \quad (12)$$

## 5. Experiments

In this section, we conduct a detailed evaluation of the proposed joint optimization framework and the BCOT

---

[1]Azure Kinect DK hardware specifications: https://docs.microsoft.com/en-us/azure/kinect-dk/hardware-specification

benchmark. We also show more results in the supplementary material. Our experimental environment is on a laptop with Intel(R) Core(TM) i7-8565U @1.8GHz CPU, NVIDIA GeForce MX250 GPU, and 8GB RAM.

### 5.1. Multi-View Tracking Evaluation

The precision of the proposed joint optimization framework is the basis for the construction of the BCOT benchmark. This section uses the synthetic data to prove that the method can obtain sufficient annotation precision. The synthetic data contains 4 objects, namely $Cat$, $Clown$, $Driller$ and $Squirrel$, and there diameters are $127.6mm$, $142.3mm$, $229.5mm$, and $194.3mm$. Each object includes 3 modes of multi-view data, i.e.: 1) *Object moves freely with fixed cameras*, 2) *Object rotates only with fixed cameras* and 3) *Cameras move freely*. Its resolution is 640×480px.

**Binocular tracking result.** We first perform the binocular tracking evaluation on the mode 1, as shown in Table 3. The basic camera C-0 and other cameras constitute binocular data. C-0 to C-8 are on an arc, forming a $plane$ with the object. C-9 to C-12 are outside the plane, and C-0/C-1/C-9, C-0/C-2/C-10, C-0/C-3/C-11 and C-0/C-4/C-12 constitute 4 sets of $cone$-type cameras with the object. The data in the table is the average error of the two views, which also averaged all 4 objects. Lost Number represents the number of tracking failure, i.e., the rotation error is larger than $5°$, or the translation error is larger than $5cm$, under the C-0 coordinate frame (first camera of the set). When the object is lost, we reset the GT pose.

The second column of Table 3 is the tracking result under C-0 with only monocular data. Overall, the rotation and translation errors of our binocular tracking gradually decrease with the included angle increasing between $5°$ to $90°$, and this tendency is especially obvious in the $Z$-axis direction. When the included angle is $90°$, the $Z$-axis translation error is $0.28mm$, which is less than 2‰ of the diameter of the model.

The object may be lost in the case that the camera included angle is small. This is because the uncertainty of the pose cannot be eliminated in a small view angle. As the included angle increases, while the object may be lost in one view, the other view can constrain the object so that the joint optimization will pull the object back to the correct pose.

**Multi-view tracking result.** We further introduce the error trend when the camera number is gradually increasing. We take C-0 as the basic camera and append cameras according to the following two patterns, i.e., the camera included angle is appended from small to large (pattern-1, C-0/C-4/C-5/C-6/C-7), and the camera included angle is appended from large to small (pattern-2, C-0/C-7/C-5/C-6/C-4). The error trends of translation and rotation are shown in Fig. 5.

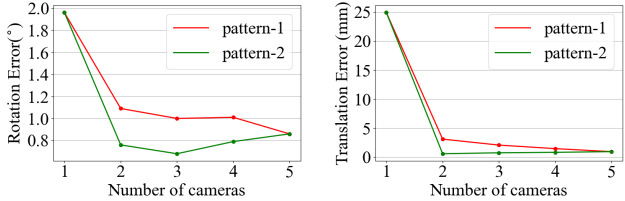

Figure 5. Multi-view tracking results. With the increase in the number of cameras, the error trend of two patterns.

| Included Angle | Mono. | Freely | Freely | Freely |
|---|---|---|---|---|
| Camera Index | C-0 | C-0/C-1 | C-0/C-2 | C-0/C-1/C-2 |
| **r**($°$) | 1.47 | 0.59 | 0.80 | **0.50** |
| tx($mm$) | 1.18 | 0.24 | 0.26 | **0.21** |
| ty($mm$) | 1.22 | 0.26 | 0.18 | **0.16** |
| **tz**($mm$) | 12.96 | 0.78 | 0.33 | **0.32** |
| Lost Number | 17 | 0 | 0 | 0 |

Table 4. Multi-view tracking evaluation on *Cameras move freely* mode.

| Reso.(width) | 320 | 640 | 1280 | 1920 | 2560 | 4096 |
|---|---|---|---|---|---|---|
| **r**($°$) | 1.01 | 0.68 | 0.40 | 0.37 | **0.34** | 0.36 |
| tx($mm$) | 0.28 | 0.21 | 0.13 | 0.10 | 0.07 | **0.06** |
| ty($mm$) | 0.40 | 0.21 | 0.11 | 0.08 | 0.05 | **0.03** |
| **tz**($mm$) | 0.28 | 0.20 | 0.12 | 0.10 | 0.08 | **0.06** |
| Lost Number | 0 | 0 | 0 | 0 | 0 | 0 |

Table 5. Binocular tracking evaluation on different resolution.

We can see that translation and rotation errors of pattern-1 show a downward trend. But in pattern-2, adding a camera between the existing cameras may increase the error, especially the translation component. Moreover, the multi-view performance in most cases is inferior to the binocular tracking with a large included angle, i.e., the green point of 2 cameras. All of these show that the camera included angle has a decisive effect on the tracking precision. The results of mode 2 are consistent with the results of mode 1. The other results can be found in the supplementary material.

**Cameras move freely.** We next evaluate the *Cameras move freely* mode. The relative transformation between the cameras is continuously changing, and we use the ground truth poses of each camera to calculate it. The average error is shown in Table 4.

We can see that when the two cameras are moving freely, our method can track precisely. The precision is slightly improved by extending to 3 cameras. This is because some small included angle situations during the tracking may increase the error, and appending cameras with larger included angle can reduce this kind of error. Overall, the free movement of the two cameras can already satisfy multi-view tracking in general scenes.

**Evaluation with different resolutions.** Table 5 shows the evaluation of the precision with different resolutions under binocular tracking. We use $Cat$ model for the evaluation, whose error is lower than the average error due to its unique structure. The camera included angle here is $90°$, and the resolution gradually increases from 320×240px to 4096×3072px. As the resolution increases, the precision of rotation and translation also gradually increases. When the resolution is 2560px, each axis's translation error is less than $0.1mm$, reaching sub-millimeter level.

## 5.2. Evaluation on BCOT Benchmark

**BCOT benchmark examples.** Fig. 6 shows examples of the BCOT benchmark. The red contour is rendered according to the annotation pose, and it can accurately align with the object contour on the image.

**Monocular 3D tracking method evaluation.** Table 6 shows the evaluation results of $n°$, $n\ cm$ and ADD metric. The $d$ in the ADD metric represents the longest side of the bounding box of the object model. For all the metrics in the table, if the rotation error is larger than $5°$ or the translation error is larger than $5cm$, we reset the GT pose. All methods use the code provided by the author. ACCV2020 [35] achieves the highest accuracy under $5°$, $5cm$. However, it needs to pre-render the template when meeting the new object, which will cost several minutes. Considering $5°$, $5cm$ is a relatively relaxed metric for tracking, we further test the accuracy under $2°$, $2cm$. We find that TVCG2021 [15] has higher rotation accuracy, and ACCV2020 [35] has higher translation accuracy.

Fig. 7 shows the tracking accuracy under different ADD error tolerances. The unit of the horizontal axis is the model side length $d$. A more detailed comparative comparison can be found in the supplementary material.

## 5.3. Limitation and Future Work

Although our multi-view joint optimization framework can achieve sufficient precision for benchmark construction, it still has some limitations. We need to keep the relationship between cameras relatively fixed for multi-view tracking, which limits many application scenarios. In future work, we will explore high-precision optimization methods with freely moving cameras. For the BCOT benchmark, to ensure precision, the movement speed of the object is relatively slow. In addition, there is currently no proper method to evaluate the rotation error of the annotation, and we will explore it in future work.

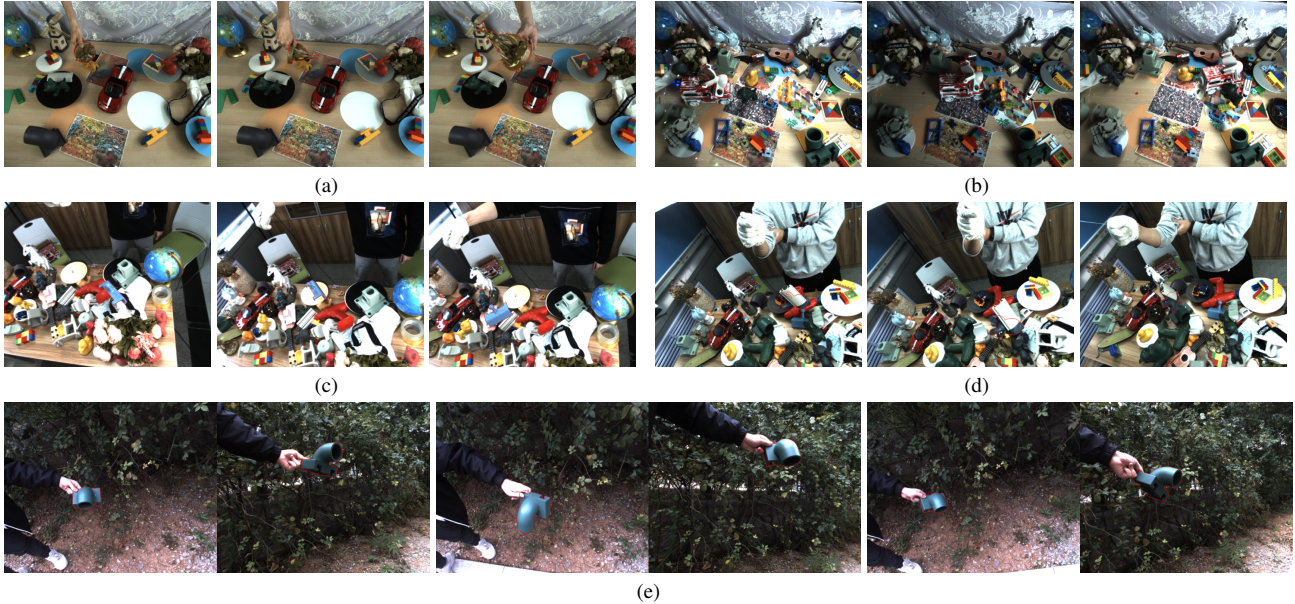

(a) (b) (c) (d) (e)

Figure 6. Examples of the BCOT benchmark, where the red contour is rendered according to the annotation pose: (a) Vampire Queen model, static camera set, easy scene, handheld movement; (b) Flashlight model, static camera set, complex scene, dynamic light, translation movement; (c) Bracket model, movable camera set, complex scene, suspension movement; (d) Deadpool model, movable camera set, complex scene, occlusion, suspension movement; (e) Standtube model, movable camera set, outdoor scene, handheld movement, providing both two views.

| Method | ADD$-0.02d$ | ADD$-0.05d$ | ADD$-0.1d$ | $5°, 5cm$ | $5°$ | $5cm$ | $2°, 2cm$ | $2°$ | $2cm$ | Time$(ms)$ |
|---|---|---|---|---|---|---|---|---|---|---|
| MTAP2019 [40] | 5.5 | 32.7 | 64.6 | 54.4 | 54.9 | 97.8 | 12.4 | 13.7 | 77.9 | 8.8 |
| TPAMI2019 [38] | 11.7 | 31.6 | 57.1 | 77.1 | 79.2 | 91.7 | 40.8 | 48.3 | 67.8 | 34.6 |
| CGF2020 [16] | 12.0 | 31.3 | 57.5 | 84.1 | 85.1 | 95.7 | 45.1 | 55.1 | 70.1 | 33.0 |
| ACCV2020 [35] | 10.9 | **45.5** | **76.9** | **89.0** | **89.3** | **99.5** | 46.0 | 49.5 | **87.8** | **3.5** |
| C&G2021 [23] | 9.1 | 31.5 | 58.1 | 82.5 | 84.7 | 95.0 | 38.5 | 47.0 | 69.9 | 18.9 |
| JCST2021 [24] | 14.4 | 38.1 | 65.7 | 87.0 | 88.1 | 97.2 | 50.2 | 57.3 | 77.2 | 38.5 |
| TVCG2021 [15] | **15.6** | 39.8 | 66.1 | 87.1 | 88.5 | 96.3 | **51.4** | **59.0** | 76.4 | 34.8 |

Table 6. Comparison of monocular 3D tracking methods.

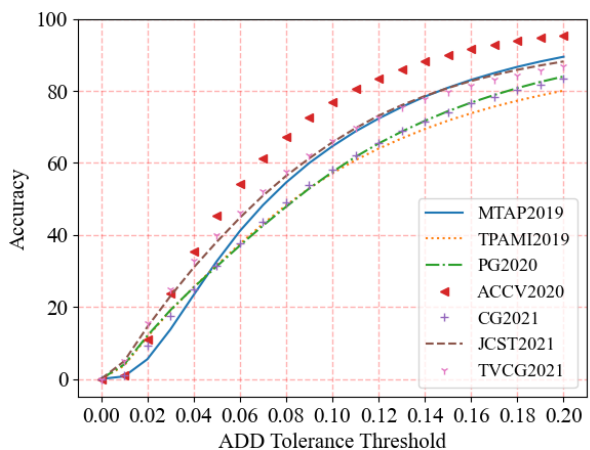

Figure 7. Overall tracking accuracy under various ADD error tolerance thresholds.

## 6. Conclusion

We have proposed a joint optimization framework for multi-view textureless 3D object tracking, based on which we further construct a real scene benchmark with high-precision annotation poses. Increasing the included angle of cameras to eliminate the pose uncertainty is the key to improving precision. The orthometric two cameras have reached enough precision, and increasing the number of cameras will not significantly improve tracking precision anymore. We comprehensively evaluated SOTA 3D object tracking methods on the proposed BCOT benchmark. At the same time, the benchmark provides real data for deep learning model training, which enables future research on tracking methods based on deep learning.

**Acknowledgements:** This work is partially supported by the National Key R&D Program of China (No. 2020YFB1708903), Zhejiang Lab (No. 2020NB0AB02), and NSF of China (No. 62172260).

# Supplementary Material of
# BCOT: A Markerless High-Precision 3D Object Tracking Benchmark

Jiachen Li[1], Bin Wang[1], Shiqiang Zhu[2], Xin Cao[1], Fan Zhong[1], Wenxuan Chen[2],
Te Li[2 *], Jason Gu[3] and Xueying Qin[1*]
[1]Shandong University    [2]Zhejiang Lab    [3]Dalhousie University

## Appendix

The appendix is organized as follows. Sec. A gives a detailed derivation of the optimization of the object-centered model and the joint framework. Sec. B presents additional test results in our synthetic multi-view dataset and shows more intermediate results. Sec. C shows more details and examples of the proposed BCOT benchmark and evaluates state-of-the-art monocular 3D tracking methods comprehensively. Finally, Sec. D compares BCOT with other 3D object tracking datasets in detail.

## A. Optimization

In this section, we first derive the Jacobian matrix in detail based on the object-centered model, based on which we solve the pose for the proposed joint optimization framework.

### A.1. Optimization of the Object-centered Model

As shown in the manuscript, we translate the camera-centered model to the object-centered model and then expand it, i.e.:

$$\boldsymbol{x} = \pi(\boldsymbol{K}(^c\boldsymbol{T}_t \tilde{\boldsymbol{X}}_t)_{3\times 1}) \tag{1}$$

$$= \pi(\boldsymbol{K}(^o\boldsymbol{T}_c^{-1} \, ^o\boldsymbol{T}_c \, ^c\boldsymbol{T}_t \tilde{\boldsymbol{X}}_t)_{3\times 1}) \tag{2}$$

$$= \pi(\boldsymbol{K}(^o\boldsymbol{T}_c^{-1} \tilde{\boldsymbol{X}}_o)_{3\times 1}) \tag{3}$$

$$= \pi\left(\boldsymbol{K}\left(\begin{bmatrix} t_{11} & t_{12} & t_{13} & t_{14} \\ t_{21} & t_{22} & t_{23} & t_{24} \\ t_{31} & t_{32} & t_{33} & t_{34} \\ 0 & 0 & 0 & 1 \end{bmatrix} \begin{bmatrix} X_o \\ Y_o \\ Z_o \\ 1 \end{bmatrix}\right)_{3\times 1}\right) \tag{4}$$

$$= \pi\left(\boldsymbol{K}\left(\begin{bmatrix} t_{11}X_o + t_{12}Y_o + t_{13}Z_o + t_{14} \\ t_{21}X_o + t_{22}Y_o + t_{23}Z_o + t_{24} \\ t_{31}X_o + t_{32}Y_o + t_{33}Z_o + t_{34} \\ 1 \end{bmatrix}\right)_{3\times 1}\right) \tag{5}$$

$$= \pi\left(\begin{bmatrix} f_x & 0 & c_x \\ 0 & f_y & c_y \\ 0 & 0 & 1 \end{bmatrix} \begin{bmatrix} A \\ B \\ C \end{bmatrix}\right) \tag{6}$$

$$= \begin{bmatrix} \frac{f_x A + c_x C}{C} \\ \frac{f_y B + c_y C}{C} \end{bmatrix}, \tag{7}$$

---

*Corresponding author: Xueying Qin (qxy@sdu.edu.cn) and Te Li (lite@zhejianglab.com)

where $A = t_{11}X_o + t_{12}Y_o + t_{13}Z_o + t_{14}$, $B = t_{21}X_o + t_{22}Y_o + t_{23}Z_o + t_{24}$, and $C = t_{31}X_o + t_{32}Y_o + t_{33}Z_o + t_{34}$.

The object-centered model and the camera-centered model are obtained from the camera projection model, which is irrelevant to the feature extraction of the tracking method. Therefore, mapping the camera-centered model to the object-centered model is universal and can be replaced in all 3D tracking methods.

The Jacobian matrix under $\boldsymbol{O}_o$ can be formulated as:

$$\boldsymbol{J}_o(\boldsymbol{x}) = \frac{\partial F}{\partial \Delta \boldsymbol{\xi}_o} = \frac{\partial F}{\partial \boldsymbol{x}} \frac{\partial \boldsymbol{x}}{\partial \boldsymbol{X}_o} \frac{\partial \boldsymbol{X}_o}{\partial \Delta \boldsymbol{\xi}_o}. \tag{8}$$

$\frac{\partial F}{\partial \boldsymbol{x}}$ is related to the energy function, which is unique to different methods and irrelevant to the coordinate frame. $\frac{\partial \boldsymbol{x}}{\partial \boldsymbol{X}_o} \frac{\partial \boldsymbol{X}_o}{\partial \Delta \boldsymbol{\xi}_o}$ is derived based on the object-centered model, which is common to each method, replacing $\frac{\partial \boldsymbol{x}}{\partial \boldsymbol{X}_c} \frac{\partial \boldsymbol{X}_c}{\partial \Delta \boldsymbol{\xi}_c}$ based on the camera-centered model.

Then we derivative $\boldsymbol{x}$ as:

$$\frac{\partial \boldsymbol{x}}{\partial \Delta \boldsymbol{\xi}_o} = \frac{\partial \boldsymbol{x}}{\partial \boldsymbol{X}_o} \frac{\partial \boldsymbol{X}_o}{\partial \Delta \boldsymbol{\xi}_o}, \tag{9}$$

where

$$\frac{\partial \boldsymbol{x}}{\partial \boldsymbol{X}_o} = \begin{bmatrix} \frac{\partial x}{\partial X_o} & \frac{\partial x}{\partial Y_o} & \frac{\partial x}{\partial Z_o} \\ \frac{\partial y}{\partial X_o} & \frac{\partial y}{\partial Y_o} & \frac{\partial y}{\partial Z_o} \end{bmatrix} \tag{10}$$

$$= \begin{bmatrix} \frac{(f_x \frac{\partial A}{\partial X_o} + c_x \frac{\partial C}{\partial X_o})C - (f_x A + c_x C)\frac{\partial C}{\partial X_o}}{C^2} & \frac{\partial x}{\partial Y_o} & \frac{\partial x}{\partial Z_o} \\ \frac{(f_y \frac{\partial B}{\partial X_o} + c_y \frac{\partial C}{\partial X_o})C - (f_y B + c_y C)\frac{\partial C}{\partial X_o}}{C^2} & \frac{\partial y}{\partial Y_o} & \frac{\partial y}{\partial Z_o} \end{bmatrix} \tag{11}$$

and

$$\frac{\partial \boldsymbol{X}_o}{\partial \Delta \boldsymbol{\xi}_o} = \frac{\partial exp(\hat{\boldsymbol{\xi}}_o) \boldsymbol{X}}{\partial \Delta \boldsymbol{\xi}_o} \tag{12}$$

$$= \lim_{\Delta \boldsymbol{\xi}_o \to \boldsymbol{0}} \frac{exp(\Delta \hat{\boldsymbol{\xi}}_o) exp(\hat{\boldsymbol{\xi}}_o) \boldsymbol{X} - \exp(\hat{\boldsymbol{\xi}}_o) \boldsymbol{X}}{\Delta \boldsymbol{\xi}_o} \tag{13}$$

$$= \lim_{\Delta \boldsymbol{\xi}_o \to \boldsymbol{0}} \frac{(\boldsymbol{I} + \Delta \hat{\boldsymbol{\xi}}_o) exp(\hat{\boldsymbol{\xi}}_o) \boldsymbol{X} - \exp(\hat{\boldsymbol{\xi}}_o) \boldsymbol{X}}{\Delta \boldsymbol{\xi}_o} \tag{14}$$

$$= \lim_{\Delta \boldsymbol{\xi}_o \to \mathbf{0}} \frac{\Delta \hat{\boldsymbol{\xi}}_o exp(\hat{\boldsymbol{\xi}}_o) \boldsymbol{X}}{\Delta \boldsymbol{\xi}_o} \quad (15)$$

$$= \lim_{\Delta \boldsymbol{\xi}_o \to \mathbf{0}} \frac{\begin{bmatrix} \Delta \hat{\boldsymbol{\phi}}_o & \Delta \boldsymbol{\rho}_o \\ \mathbf{0}^\top & 0 \end{bmatrix} \begin{bmatrix} \boldsymbol{RX} + \boldsymbol{t} \\ 1 \end{bmatrix}}{\Delta \boldsymbol{\xi}_o} \quad (16)$$

$$= \lim_{\Delta \boldsymbol{\xi}_o \to \mathbf{0}} \frac{\begin{bmatrix} \Delta \hat{\boldsymbol{\phi}}_o (\boldsymbol{RX} + \boldsymbol{t}) + \Delta \boldsymbol{\rho}_o \\ 0 \end{bmatrix}}{\Delta \boldsymbol{\xi}_o} \quad (17)$$

$$= \begin{bmatrix} \boldsymbol{I} & -(\boldsymbol{RX} + \boldsymbol{t})^\wedge \\ \mathbf{0}^\top & \mathbf{0}^\top \end{bmatrix} \quad (18)$$

$$= [\boldsymbol{I} \quad -\hat{\boldsymbol{X}}_o]. \quad (19)$$

$\boldsymbol{\phi}_o$ and $\boldsymbol{\rho}_o$ are the rotation and translation components of $\boldsymbol{\xi}_o$, and $\boldsymbol{R}$ and $\boldsymbol{t}$ are the corresponding rotation matrix and translation vector of $\boldsymbol{T}$.

### A.2. Optimization of the Joint Framework

For the $i$-th camera, we can calculate the Jacobian matrix of the object-centered model as described above, i.e.:

$$\boldsymbol{J}_o^i(\boldsymbol{x}) = \frac{\partial F^i}{\partial \Delta \boldsymbol{\xi}_o} = \frac{\partial F^i}{\partial \boldsymbol{x}} \frac{\partial \boldsymbol{x}}{\partial \boldsymbol{X}_o} \frac{\partial \boldsymbol{X}_o}{\partial \Delta \boldsymbol{\xi}_o}. \quad (20)$$

We use the Gauss-Newton method for optimization, where the second-order Taylor approximation of the energy function $E$ is formulated as:

$$E(\boldsymbol{\xi}_o + \Delta \boldsymbol{\xi}_o) \approx E(\boldsymbol{\xi}_o) + \sum_{i=1}^N \sum_{\boldsymbol{x} \in \Omega^i} \boldsymbol{J}_o^i(\boldsymbol{x}) \Delta \boldsymbol{\xi}_o \\ + \frac{1}{2} \sum_{i=1}^N \sum_{\boldsymbol{x} \in \Omega^i} \Delta \boldsymbol{\xi}_o^\top \boldsymbol{J}_o^{i\top}(\boldsymbol{x}) \boldsymbol{J}_o^i(\boldsymbol{x}) \Delta \boldsymbol{\xi}_o. \quad (21)$$

In Eq. 21, the second-order derivative is dropped when calculating the Hessian matrix. Then the update step in the object-centered model can be formulated as:

$$\Delta \boldsymbol{\xi}_o = -\Big(\sum_{i=1}^N \sum_{\boldsymbol{x} \in \Omega^i} \boldsymbol{J}_o^{i\top}(\boldsymbol{x}) \boldsymbol{J}_o^i(\boldsymbol{x})\Big)^{-1} \\ \cdot \sum_{i=1}^N \sum_{\boldsymbol{x} \in \Omega^i} \boldsymbol{J}_o^{i\top}(\boldsymbol{x}). \quad (22)$$

Finally, we map $\Delta \boldsymbol{\xi}_o$ to each camera coordinate frame, i.e.:

$$\Delta \boldsymbol{T}^i = {}^{c_i}\boldsymbol{T}_o exp(\Delta \hat{\boldsymbol{\xi}}_o)({}^{c_i}\boldsymbol{T}_o)^{-1}. \quad (23)$$

## B. Additional Multi-view Tracking Results on the Synthetic Data

The synthetic dataset contains three modes of multi-view data, including: *1) Object moves freely with fixed cameras*, *2) Object rotates only with fixed cameras* and *3) Cameras move freely*.

In the manuscript, we give the results of binocular tracking and multi-view tracking results in mode 1. In this section, we first provide the rest results in mode1 and mode 2 and then give some intermediate results to show the effectiveness of our method. Fig. 1 shows the spatial distribution of cameras. The red cameras constitute the *plane*-type, and the green cameras with the other two cameras on the plane constitute the *cone*-type.

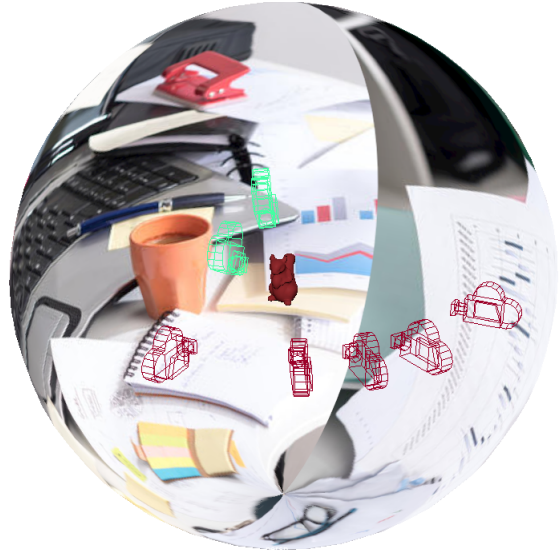

Figure 1. Spatial distribution of cameras in first two modes. The red cameras and the object constitute a *plane*, while the green cameras are outside the plane, constituting the *cone* with the other two cameras on the plane.

### B.1. Trinocular Tracking Result in Mode 1

We select several groups of cameras for the trinocular tracking evaluation in mode 1. The selection principle is that the included angle between the first camera and the second camera is equal to the included angle between the second camera and the third camera. Based on this, we select 8 sets of data where the first four are the object and the cameras constitute a *plane*, and the last four are the object and the cameras constitute a *cone*. The camera angles are $5°$, $10°$, $30°$, and $45°$, respectively. Table 1 shows the evaluation results.

Overall, the rotation and translation errors decrease with the camera angle increase in both *plane* and *cone* cameras. We can find more interesting consequences by combining Table 3 in the manuscript and Table 1, that is, 1) when the object and the cameras on one *plane*, the precision depends on the two cameras with the largest included angle, and adding cameras between them will reduce the overall precision. 2) The precision of the *cone* cameras is worse than that of the *plane* cameras if the included angle is the same.

The reason for this phenomenon is that the cameras with a large included angle can eliminate uncertainty, but adding a camera between them actually introduces the new uncer-tainty in the view direction, resulting in increased error. Therefore, in practical applications, our primary purpose is to eliminate the uncertainty of the object pose through different views, meaning that increasing the camera angle is more preferred than increasing the camera number.

| Camera Angle | Mono. | 5° plane | 10° plane | 30° plane | 45° plane | 5° cone | 10° cone | 30° cone | 45° cone |
|---|---|---|---|---|---|---|---|---|---|
| Camera Index | C-0 | C-0/C-1/C2 | C-0/C-2/C3 | C-0/C-4/C6 | C-0/C-5/C-7 | C-0/C-1/C9 | C-0/C-2/C10 | C-0/C-4/C11 | C-0/C-5/C-12 |
| **r**(°) | 1.62 | 1.25 | 1.21 | 0.93 | **0.68** | 1.28 | 1.24 | 0.96 | 0.77 |
| tx(mm) | 4.36 | 2.51 | 1.37 | 0.48 | **0.36** | 2.97 | 1.77 | 0.52 | 0.40 |
| ty(mm) | 2.39 | 1.37 | 0.71 | **0.27** | **0.27** | 1.68 | 1.02 | 0.45 | 0.38 |
| **tz**(mm) | 22.09 | 12.70 | 6.58 | 1.06 | **0.45** | 15.26 | 9.04 | 1.74 | 0.85 |
| Lost Number | 21 | 10 | 5 | 0 | 0 | 14 | 9 | 0 | 0 |

Table 1. Trinocular tracking evaluation on *Object moves freely with fixed cameras* mode.

| Camera Angle | Mono. | 30° | 45° | 60° | 90° | 30° | 45° |
|---|---|---|---|---|---|---|---|
| Camera Index | C-0 | C-0/C-1 | C-0/C-2 | C-0/C-3 | C-0/C-4 | C-0/C-5 | C-0/C-6 |
| **r**(°) | 1.41 | 1.15 | 1.05 | 1.06 | **0.79** | 0.93 | 0.87 |
| tx(mm) | 0.39 | 0.64 | 0.55 | 0.38 | 0.36 | **0.22** | 0.31 |
| ty(mm) | 0.37 | 0.13 | 0.13 | **0.12** | 0.13 | 0.45 | 0.42 |
| **tz**(mm) | 15.07 | 2.48 | 1.37 | 0.61 | **0.36** | 1.92 | 1.39 |
| Lost Number | 9 | 0 | 0 | 0 | 0 | 0 | 0 |

Table 2. Binocular tracking evaluation on *Object rotates only with fixed cameras* mode.

| Camera Angle | Mono. | 30° plane | 45° plane | 30° cone | 45° cone |
|---|---|---|---|---|---|
| Camera Index | C-0 | C-0/C-1/C-3 | C-0/C-2/C-4 | C-0/C-1/C-5 | C-0/C-2/C-6 |
| **r**(°) | 1.41 | 1.12 | 0.86 | 0.93 | **0.85** |
| tx(mm) | 0.39 | 0.39 | 0.39 | 0.34 | **0.31** |
| ty(mm) | 0.37 | **0.12** | **0.12** | 0.31 | 0.28 |
| tz(mm) | 15.07 | 0.90 | **0.60** | 1.55 | 0.99 |
| Lost Number | 9 | 0 | 0 | 0 | 0 |

Table 3. Trinocular tracking evaluation on *Object rotates only with fixed cameras* mode.

If we use multiple cameras, we should also increase the included angle between them as much as possible, uniformly distributed in space.

### B.2. Binocular Tracking Results in Mode 2

Table 2 shows the binocular tracking evaluation results in mode 2, i.e., *Object rotates only with fixed cameras*. In this mode, C-0 to C-4 and the object constitute a *plane*, and C-5 and C-6 are outside the plane. Since the object only rotates, the translation errors in the $X$-axis and $Y$-axis directions during tracking are tiny. The $X$-axis translation error of some binocular tracking is larger than that of monocular tracking, which is caused by the geometric shape of the object.

Generally, when the camera angle is within 90°, as the camera angle increases, the rotation and translation errors gradually decrease, especially the translation in the $Z$-axis direction, which is consistent with the conclusions in the manuscript.

Since C-5 and C-6 are looking at the object from a higher position, the translation components of the $X$-axis and the $Y$-axis are a little different from C-1 to C-4, where the $X$-axis component is better but $Y$-axis component is worse.

### B.3. Trinocular Tracking Results in Mode 2

Table 3 gives the trinocular evaluation results under mode 2. For the translation component, the *plane* pattern is better than the *cone* pattern, i.e., the precision depends on the two cameras with the largest angle. For the rotation component, the precision of the two patterns is very close, which is caused by the object only rotating.

Combining Table 2 and Table 3, we find that adding cameras between the two cameras will reduce the tracking accuracy, which is consistent with the conclusions above. For example, the $Z$-axis translation precision of C-0/C-4 is $0.36mm$, while the corresponding precision in C-0/C-2/C-4 is only $0.60mm$. At the same time, the rotation precision is also reduced.

### B.4. Intermediate Results

This section analyzes the detailed effect of our multi-view method through the intermediate results. Fig. 2 shows the binocular tracking results results on real data. The first row is the input images of the two cameras with 90° included angle, the second row is the tracking result of TPAMI19, which performs the monocular tracking, and the third row is our multi-view tracking result. The C-1 image is the result of rendering with $T_1$, and it can be seen that the reprojection region of TPAMI19 can precisely match the input image visually. The C-2 image is the rendering result of $T'_2$, that is, transforming $O_{c_1}$ to $O_{c_2}$ by $T'_2 =^2T_1T_1$ for rendering. We can see that TPAMI19 has an obvious translation error in the camera view direction. Our joint optimization can get the precise pose, resulting in the reprojection region on each camera image is precise. Fig. 3 shows the intermediate results on the synthetic data (*Cameras move freely* mode), which is consist with Fig. 2.

Fig. 4 is another set of results on the synthetic data. We use two cameras with 90° included angle to estimate the object pose and then observe the object with three other views, i.e., 30°, 45°, and 60°. We enlarge the image for better observation, where the purple contour in the figure is the rendering result, and we can see that they can be visually aligned with the object contour precisely in all views.

Fig. 5 is the trinocular tracking result on real data, which is also enlarged for better observation. We can see that our method can get excellent visual performances under all views.

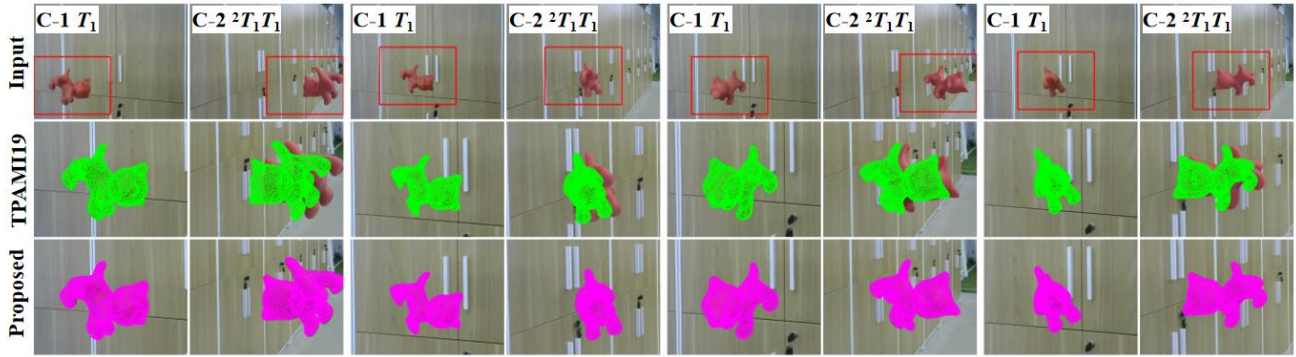

Figure 2. Binocular tracking results results on the real data. There are 4 sets of images captured from C-1 and C-2. TPAMI19 perform the monocular tracking on C-1 and map it to C-2 for display through $^2T_1T_1$, where we can see a large translation error in the camera view direction. Our method can joint the information of two cameras to get precise visual results under each camera.

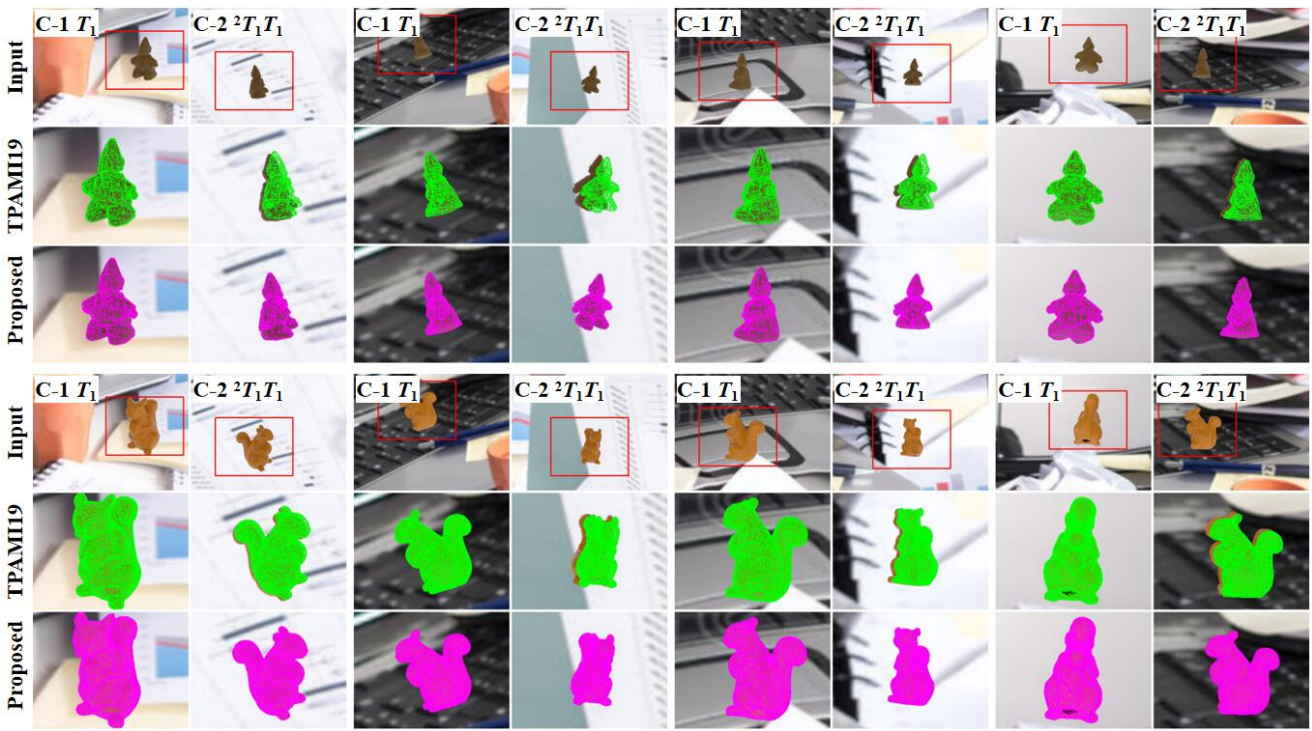

Figure 3. Binocular tracking results on the synthetic data (*cameras freely move*).

## C. Monocular Tracking Evaluation on the BCOT Benchmark

In this section, we will give more details of the BCOT benchmark. Then we show more examples and evaluate state-of-the-art monocular tracking algorithms comprehensively.

### C.1. More Details of the BCOT Benchmark

**Time cost and iterations.** The tracking time of the proposed multi-view tracking method depends on the selected basic monocular tracking method and the number of cameras used. There is no extra time needed to convert the basic coordinate frame to the object-centered coordinate frame.

For pose annotation when constructing BCOT Benchmark, the optimization executes 5 rounds (7 iterations per round), while for normal cases, only 1 round is required.

**Sequences discarded.** As stated in the manuscript, we will discard sequences with large errors. Specifically, we discarded 36 sequences, i.e., 20×22-404=36.

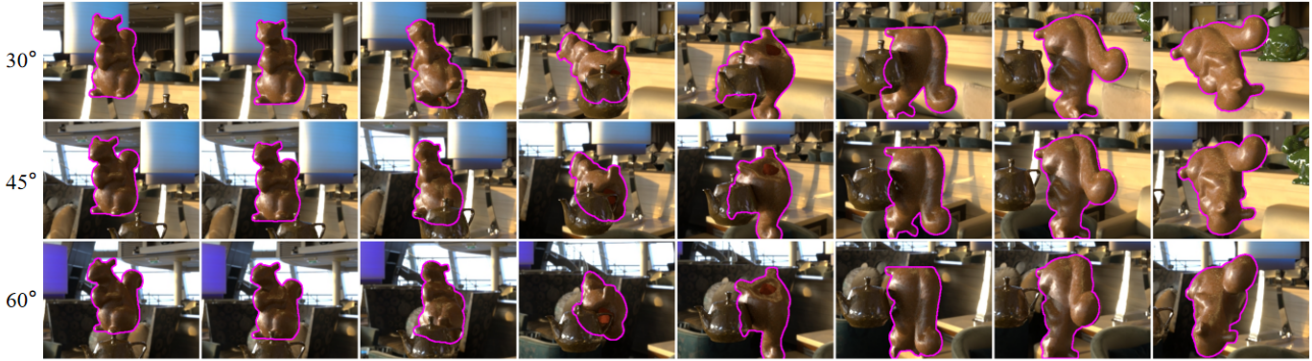

Figure 4. We use the basic camera (0° camera) and 90° camera to estimate the object pose and then observe the tracking result with three other perspectives, i.e., 30°, 45°, and 60° cameras. The object is precisely aligned with the image in all views.

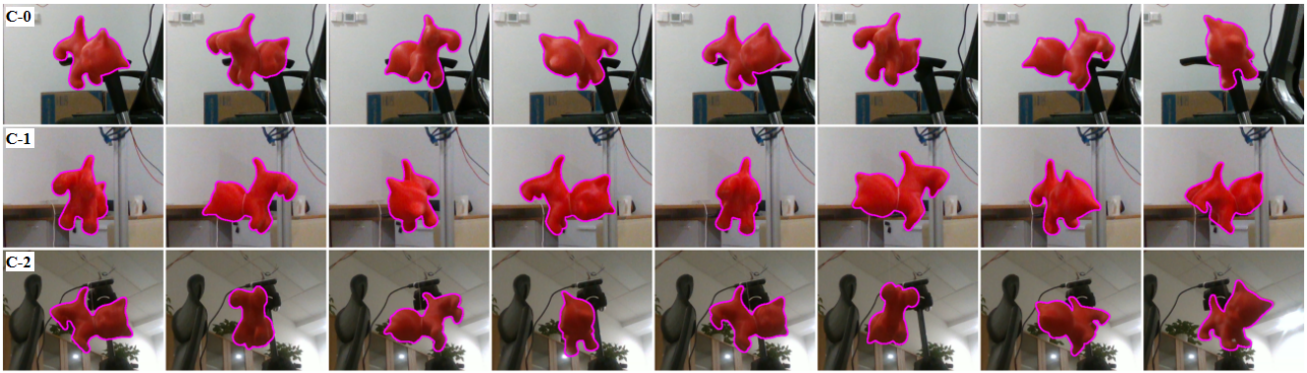

Figure 5. Trinocular tracking results on the real data. C-0 to C-2 represent the images captured by 3 cameras, respectively, and our method can get precise results in each view.

### C.2. More Examples of the BCOT Benchmark

Fig. 6−8 shows more BCOT benchmark examples. The blue contour is the result rendered according to the annotation pose. Our benchmark is markerless and can annotate high-precision poses.

### C.3. Monocular Tracking Evaluation

We further analyze the performance of the methods in indoor and outdoor scenes. Table 4 and Table 5 respectively show the accuracy of different methods in indoor and outdoor scenes. Fig. 9 and Fig. 10 respectively show the AUC scores of the ADD metric in the indoor scene and the outdoor scene.

**Analysis.** Through comparative analysis, the ACCV2020 gets the best performance overall. But as stated in the manuscript, it has some limitations when directly used. Besides, the origin of the object model coordinate frame needs to be set at the center of the model, which may limit the application scenario.

With further analysis, it can be seen from Table 4 that in indoor scenes, TVCG2021 achieves the best performance in ADD metric and $2°,2cm$ metric. This indicates that TVCG2021 has higher tracking precision in complex scenes, where the complexity of indoor scenes in the BCOT dataset is higher than that of outdoor scenes. In outdoor scenes, the background is relatively simple, ACCV2020 shows the best tracking performance, but TVCG2021 still has the highest rotation precision.

The reason is ACCV2020 uses prerendered templates in fixed discrete view angles (in order for acceleration), which introduces angular errors in templates and reduces its precision in rotation estimation. On the contrary, TVCG2021 render templates online, and there is no error in templates. The translation is insensitive to the small angular error of templates and thus is less affected.

In addition, it is found from Fig. 10 that MTAP2019 shows a high AUC score under the ADD metric. This is because MTAP2019 can obtain high translation precision in a simple background (also shown in the $2cm$ metric in Table 5), and the ADD error depends more on the translation error. However, except for ACCV2020 and MTAP2019, the

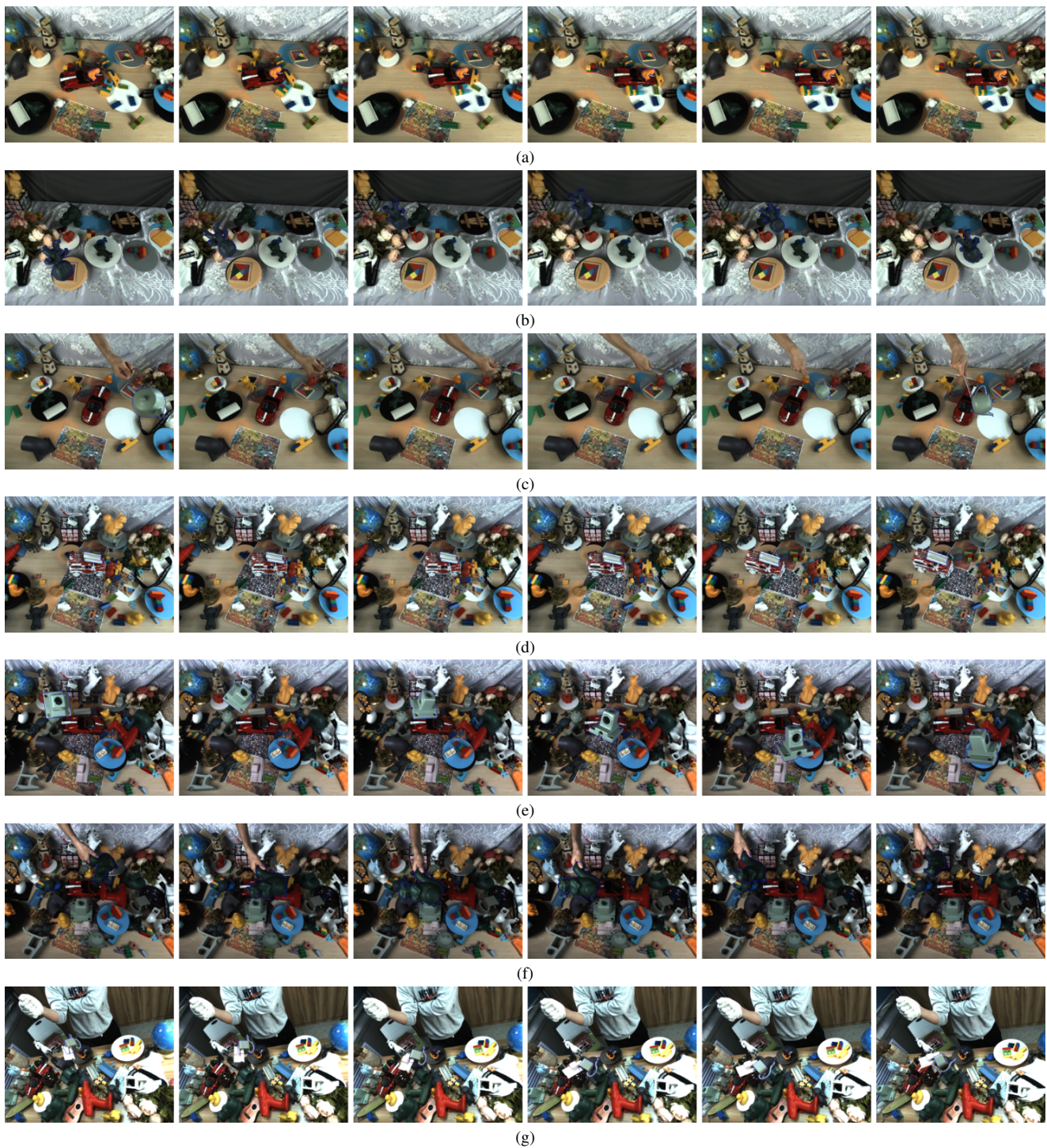

Figure 6. More examples of the BCOT benchmark, where the blue contour is rendered according to the annotation pose. (a) Ape model, static camera set, easy scene, translation movement. (b) Deadpool model, static camera set, easy scene, suspension movement. (c) Teapot model, static camera set, easy scene, handheld movement. (d) RTI Arm model, static camera set, complex scene, translation movement. (e) Lamp Clamp model, static camera set, complex scene, suspension movement. (f) Squirrel model, static camera set, complex scene, handheld movement. (g) RJ45 Clip model, movable camera set, complex scene, suspension movement, occlusion.

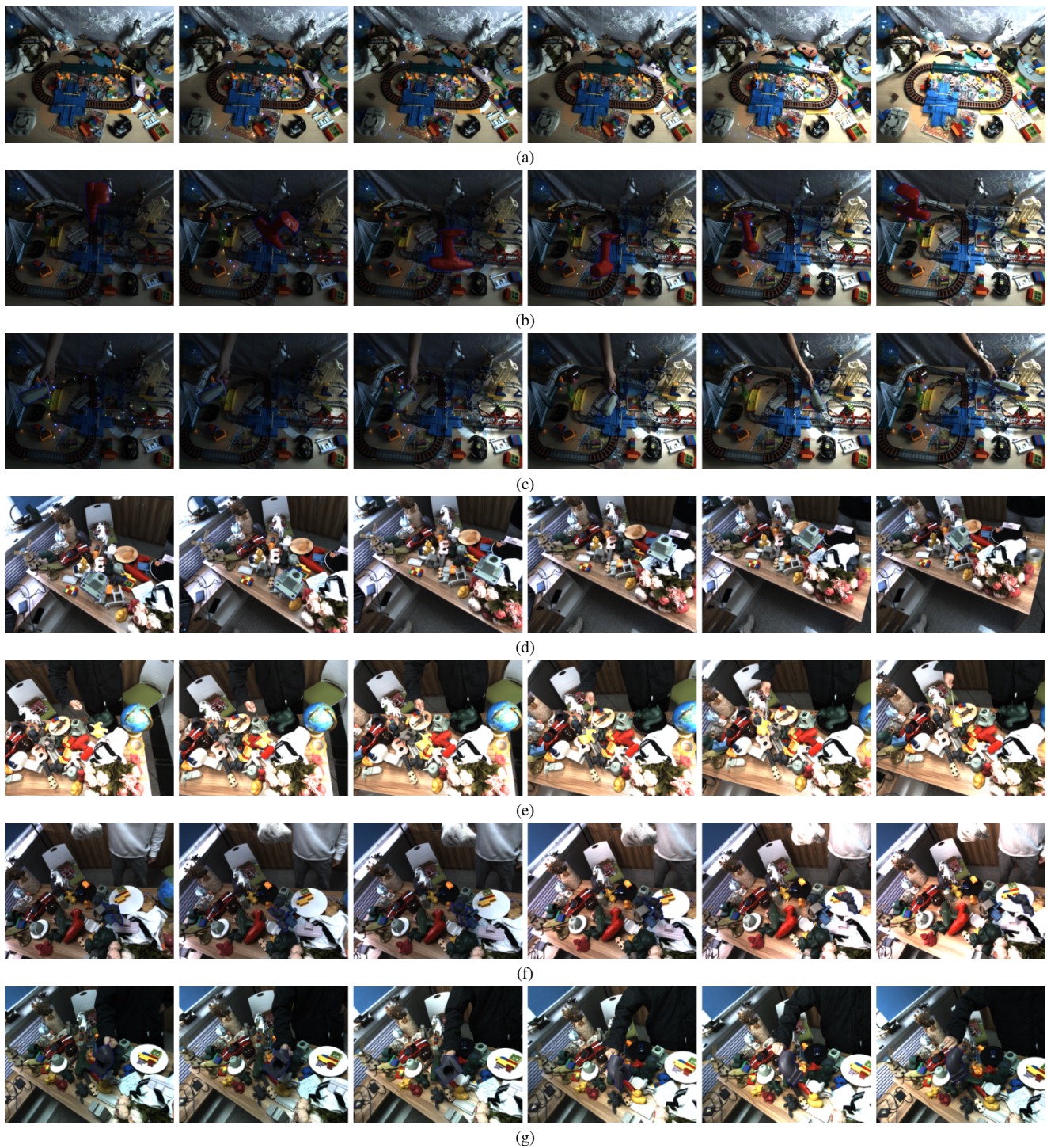

Figure 7. More examples of the BCOT benchmark, where the blue contour is rendered according to the annotation pose. (a) Wall Shelf model, static camera set, dynamic light, translation movement. (b) Driller model, static camera set, dynamic light, suspension movement. (c) 3D Touch model, static camera set, dynamic light, handheld movement. (d) Lamp Clamp model, movable camera set, complex scene, suspension movement. (e) Cat model, movable camera set, complex scene, handheld movement. (f) Stitch model, movable camera set, complex scene, dynamic light, suspension movement. (g) Tube model, movable camera set, complex scene, dynamic light, handheld movement.

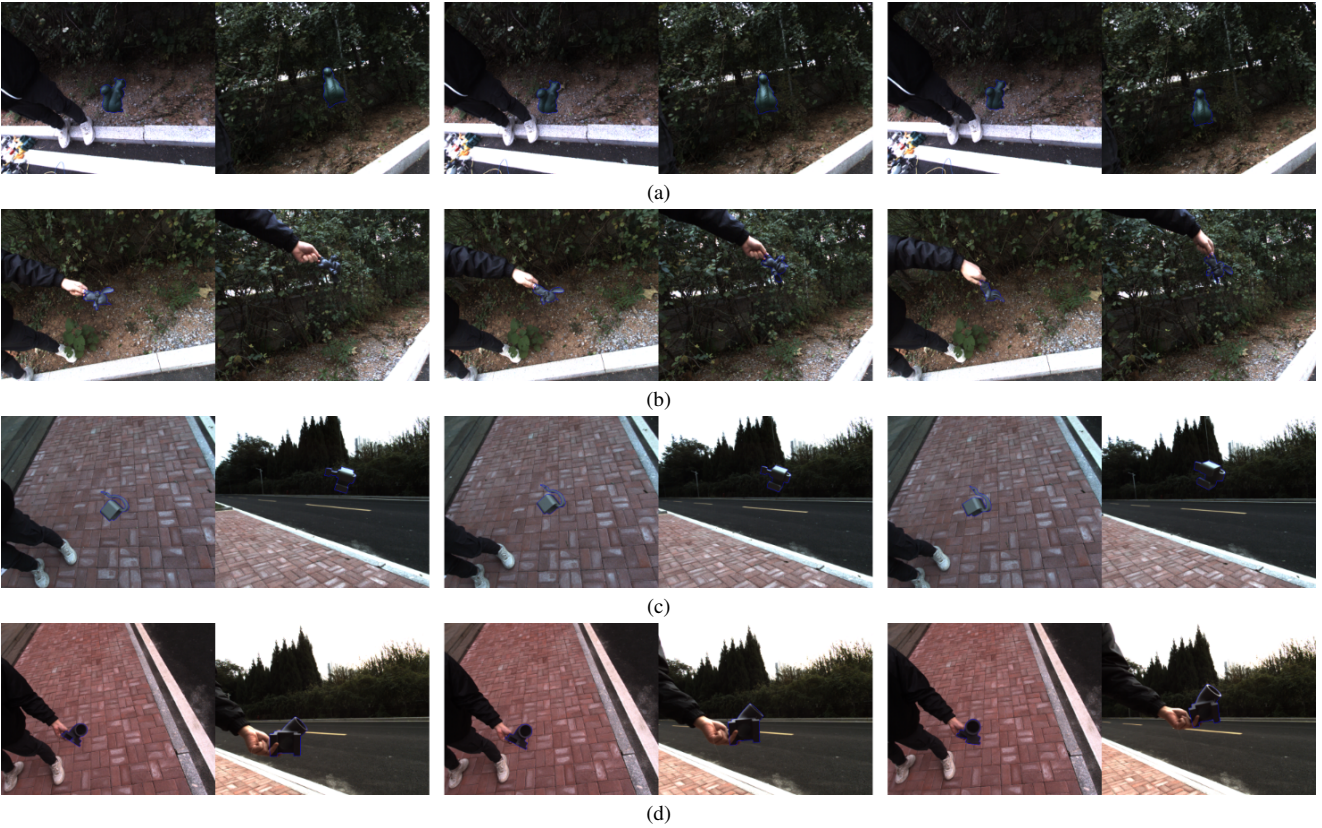

Figure 8. More examples of the BCOT benchmark, where the blue contour is rendered according to the annotation pose. The outdoor scenes provide both two views. (a) Squirrel model, outdoor scene 1, movable camera set, suspension movement. (b) Stitch model, outdoor scene 1, movable camera set, handheld movement. (c) RJ45 Clip model, outdoor scene 2, movable camera set, suspension movement. (d) Tube model, outdoor scene 2, movable camera set, handheld movement.

| Method | ADD−0.02d | ADD−0.05d | ADD−0.1d | 5°, 5cm | 5° | 5cm | 2°, 2cm | 2° | 2cm | Time(ms) |
|---|---|---|---|---|---|---|---|---|---|---|
| MTAP2019 | 7.0 | 32.5 | 61.6 | 57.9 | 58.3 | 97.8 | 15.4 | 16.9 | 75.1 | 8.6 |
| TPAMI2019 | 17.7 | 43.5 | 66.5 | 75.0 | 75.8 | 92.0 | 44.8 | 47.5 | 76.0 | 33.3 |
| CGF2020 | 18.7 | 45.5 | 70.6 | 83.2 | 83.5 | 96.0 | 52.9 | 56.8 | 81.2 | 32.2 |
| ACCV2020 | 9.8 | 43.1 | 76.4 | **88.2** | **88.4** | **99.6** | 45.6 | 49.2 | **87.8** | **3.5** |
| C&G2021 | 13.5 | 42.6 | 69.2 | 83.1 | 84.0 | 96.6 | 43.8 | 47.6 | 79.9 | 19.2 |
| JCST2021 | 21.7 | 52.0 | 76.9 | 87.0 | 87.4 | 97.8 | 55.8 | 58.5 | 86.0 | 38.8 |
| TVCG2021 | **23.6** | **55.5** | **78.5** | 87.1 | 87.4 | 97.3 | **58.2** | **60.6** | 87.1 | 32.8 |

Table 4. Comparison of monocular 3D tracking methods of indoor scenes.

| Method | ADD−0.02d | ADD−0.05d | ADD−0.1d | 5°, 5cm | 5° | 5cm | 2°, 2cm | 2° | 2cm | Time(ms) |
|---|---|---|---|---|---|---|---|---|---|---|
| MTAP2019 | 3.2 | 32.9 | 69.0 | 49.3 | 49.9 | 97.9 | 8.0 | 9.0 | 82.0 | 9.1 |
| TPAMI2019 | 2.9 | 14.0 | 43.3 | 80.3 | 84.3 | 91.4 | 34.8 | 49.5 | 55.7 | 36.6 |
| CGF2020 | 2.1 | 10.3 | 38.1 | 85.4 | 87.5 | 95.4 | 33.6 | 52.6 | 53.9 | 34.3 |
| ACCV2020 | **12.5** | **49.0** | **77.6** | **90.3** | **90.7** | **99.4** | **46.5** | 50.0 | **87.8** | **3.3** |
| C&G2021 | 2.7 | 15.2 | 41.8 | 81.7 | 85.8 | 92.7 | 30.8 | 46.1 | 55.1 | 18.4 |
| JCST2021 | 3.8 | 17.6 | 49.2 | 87.1 | 89.1 | 96.3 | 41.9 | 55.7 | 64.2 | 37.9 |
| TVCG2021 | 3.9 | 16.8 | 47.9 | 87.2 | 90.3 | 94.7 | 41.3 | **56.7** | 60.7 | 37.8 |

Table 5. Comparison of monocular 3D tracking methods of outdoor scenes.

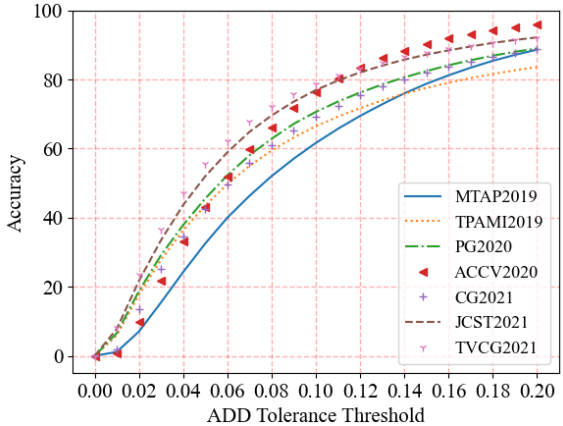

Figure 9. Indoor scene tracking accuracy under various ADD error tolerance thresholds.

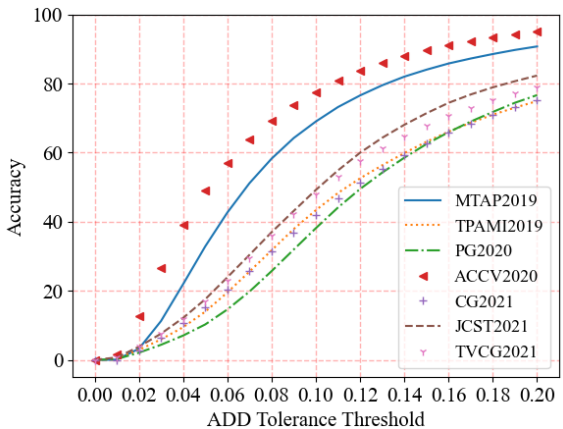

Figure 10. Outdoor scene tracking accuracy under various ADD error tolerance thresholds.

AUC scores of ADD in other methods have decreased in outdoor scenes, indicating that their translation errors have increased. This shows domain differences between outdoor and indoor scenes, which may affect the tracking method.

## D. Comparison with Other Tracking Datasets

Other datasets used for 3D object tracking include RBOT, OPT, and YCB-Video. Fig. 11 shows some examples of other datasets.

**Datas.** Fig. 11(a) is an example of the RBOT dataset. It is semi-synthetic with rendered foreground objects. The synthesized dynamic light, noise, occlusion, and object motion are very different from real scenes, which also prevents them from being used by the learning-based methods.

Fig. 11(b) is an example of the OPT dataset. It is a real scene dataset, and the GT pose is calculated by artificial markers. Objects in the dataset are always stationary, surrounded by white areas and a large number of artificial markers.

Fig. 11(c) shows an example of the YCB-Video dataset. It is also a real scene dataset, which calculates the GT pose of objects through depth data. As stated in the manuscript, this annotation method will suffer from large errors. In addition, objects in YCB-Video also remain stationary.

The main feature of BCOT is real-scene and high-precision. It is markerless, and the camera and object are both dynamic. Besides, its labeling error also achieves the highest precision. Currently, the learning-based and the optimization-based methods are studied separately on different benchmarks (RBOT v.s. YCB-Video). We believe that one important reason is the lacking of high-precision real-scene datasets, which now can be addressed with BCOT.

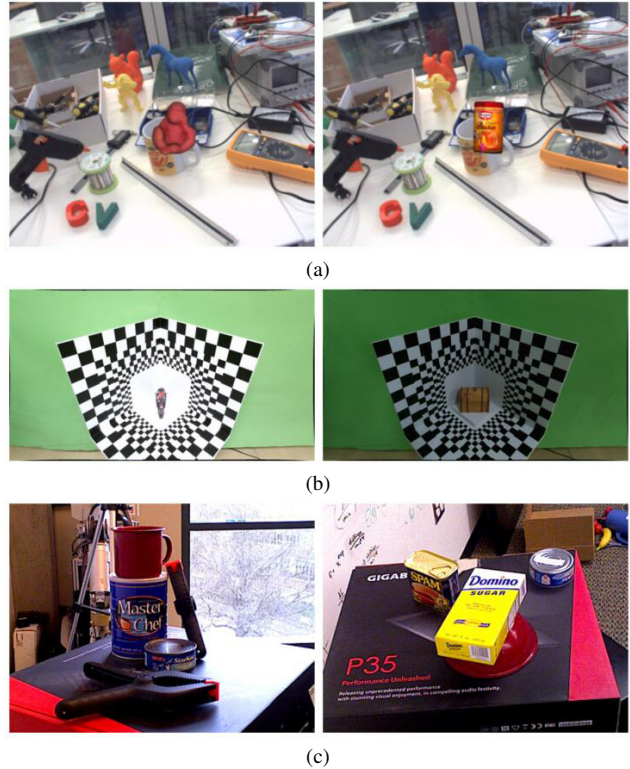

Figure 11. Examples of other datasets. (a) RBOT dataset. (b) OPT dataset. (c) YCB-Video dataset.